\documentclass[runningheads]{llncs}
\usepackage[T1]{fontenc}
\usepackage[table,xcdraw]{xcolor}
\usepackage{graphicx}
\usepackage{svg}
\usepackage{subcaption}
\usepackage{color}
\usepackage{authblk}

\begin{document}
\title{Scalable and low-cost remote lab platforms: Teaching industrial robotics using open-source tools and understanding its social implications\thanks{Accepted at Springer's 16th International Conference on Social Robotics + AI 2024}}
\titlerunning{Scalable and low-cost remote lab platforms}
%

\author[1]{Amit Kumar\orcidID{0000-0001-6698-0855}}
\author[2]{Jaison Jose\orcidID{0000-0002-4132-1359}}
\author[2]{Archit Jain\orcidID{0009-0007-4290-9537}}
\author[2]{Siddharth Kulkarni\orcidID{0009-0000-4353-3884}}
\author[3]{Kavi Arya\orcidID{0000-0002-7601-317X}}
\affil[1]{Centre for Systems and Control, Indian Institute of Technology Bombay, Mumbai, India, \email{amit.k.kumar@iitb.ac.in}}
\affil[2]{Embedded Real-Time Systems {/} e-Yantra Lab, Indian Institute of Technology Bombay, Mumbai, India, \email{\{jaisonjose241,arrchit.jain,siddharth.ydk\}@gmail.com}}
\affil[3]{Computer Science and Engineering Department, Indian Institute of Technology Bombay, Mumbai, India, \email{kavi@cse.iitb.ac.in}}
\authorrunning{A. Kumar et al.}
\institute{}
%

%
\maketitle              
\begin{abstract}
With recent advancements in industrial robots, educating students in new technologies and preparing them for the future is imperative. However, access to industrial robots for teaching poses challenges, such as the high cost of acquiring these robots, the safety of the operator and the robot, and complicated training material. This paper proposes two low-cost platforms built using open-source tools like Robot Operating System (ROS) and its latest version ROS 2 to help students learn and test algorithms on remotely connected industrial robots. Universal Robotics (UR5) arm and a custom mobile rover were deployed in different life-size testbeds, a greenhouse, and a warehouse to create an Autonomous Agricultural Harvester System (AAHS) and an Autonomous Warehouse Management System (AWMS). These platforms were deployed for a period of 7 months and were tested for their efficacy with 1,433 and 1,312 students, respectively. The hardware used in AAHS and AWMS was controlled remotely for 160 and 355 hours, respectively, by students over a period of 3 months.

\vspace{0.5cm}


\end{abstract}


\section{Introduction and Related Work}
\label{sec:intro}

One of the key aspects of Industry 5.0 is to adopt a human-centric approach to digital technologies~\cite{indus5}. With the world transitioning to this fifth industrial revolution, training students in current and upcoming technologies is necessary. However, access to the latest technologies, like those used in industrial robotics, is expensive. As a result, only a handful of students tend to learn them.
\newpage
Most manufacturers of industrial robots provide a model of their robots that can be imported into various simulators~\cite{VrepGazeCompa,GazUniCompa}.  However, in general cases, a sim-to-real~\cite{sim2realGap} gap is prevalent. Hence, it is a well-adapted practice to develop rudimentary algorithms in simulation and then make them more robust by testing them in hardware.

ROS~\cite{rosPaper} and ROS 2~\cite{ros2Paper} provide a robust modular platform for easy and fast deployment of robotic applications. They are open-source; hence, many existing libraries, in the form of ROS packages, can be configured for a particular application instead of being rewritten. Many industries and organizations like NASA~\cite{spaceROS} have also harnessed the power of ROS in their projects. It is then only logical to train students in industrial robotics using ROS.    

\begin{figure}[!t]
    \centering
    \includegraphics[width=\textwidth]{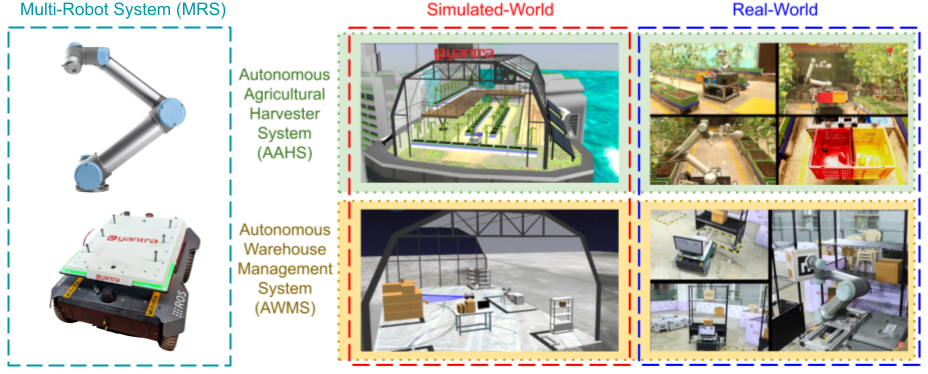}
    \caption{Set up of remote testbeds in a greenhouse (top) and a warehouse (bottom) for Autonomous Agricultural Harvester System (AAHS) and Autonomous Warehouse Management System (AWMS) using a UR5 robotic arm and a mobile rover. Students used our platforms to test their algorithms in the simulated-world (left) and real-world (right).}
    \label{fig:aahs_awms_intro}
\end{figure}

Setting up designated laboratories to provide remote access to industrial robots is not unheard of~\cite{ROS_ICRA_2015,whyRL2,whyRL3,whyRL1}. These remote labs bridge the gap between buying multiple expensive robots and giving access to many students. Several works have proposed and tested different architectures for teaching students through remote robotic labs. 

Researchers in~\cite{ORWL} have created a web app and integrated virtual network computing (VNC)-based graphical desktop-sharing and docker containers. The web app allows the users to access the small Robotont mobile robot and the UFACTORY xArm7 manipulator robot individually. They have deployed their robots in a simple and small arena, which may not reproduce all the complications generally accompanied while deploying industrial robots in real-world settings like noise in sensors due to direct light, heat, etc.

A low-cost and lightweight robotic manipulator using Raspberry Pi 3 computer is deployed in~\cite{rpiLowCost} to teach primary-school and university-level students. Although the cost of the robotic arm is low, these arms do not offer industrial standard functionalities like force-torque sensing. The payload and reach of these arms are also unsuitable to be deployed in a real-world setting. ROS was not integrated into their architecture since they aimed to introduce basic manipulation concepts to young students.

Researchers in~\cite{ROS_ICRA_2015} transfer the script to the NAO humanoid robot initially and then run the script locally. However, the user cannot obtain live feedback during the execution of the code. KUKA Robot Learning Lab~\cite{RE_Scale_ICRA} uses a queue-based parallel processing architecture for running code onto static robotic arms. Whenever a user submits a code using their web interface, it is first tested in a simulator, Gazebo, and then sent to the actual robot. Although this improves the safety of the hardware, it comes with a computational overhead. Moreover, in a real-world setting, as shown in AAHS and AWMS, where a depth camera, robotic arm, and a mobile robot work in parallel, this can significantly slow the execution. Integrating the safety of the robots in the hardware itself may be better suited for scalable applications. Additionally, none of the remote robotic labs mentioned above have tested their architecture on a large scale. These methods may not function as intended or even be expensive when thousands of students use them.

In totality, developing efficient platforms, encompassing simulations, for remote labs using open-source tools like ROS is an important and scalable step to prepare our students for Industry 5.0. One of the learning-based methodologies, project-based learning (PBL), has shown to be an effective method for learners to master new concepts, including ROS~\cite{educonSweden,PBLROS,kaEyantra,PBLAustralia,apEyantra,ICALT_Abhi}. Our research focuses on developing these remote lab platforms and testing their efficacy by deploying them in our PBL-based competition called the e-Yantra Robotics Competition (eYRC). Our platforms were used by students to develop algorithms for an Autonomous Agricultural Harvester System (AAHS) and Autonomous Warehouse Management System (AWMS), as shown in Figure~\ref{fig:aahs_awms_intro}. These platforms helped students learn about image processing, navigation algorithms for mobile rovers, motion planning algorithms for manipulators, Proportional Integral Derivative (PID) controllers, and ROS.
The major contributions of this work are as follows:
\begin{enumerate}
    \item Design of two low-cost, scalable, and customizable platforms using open-source tools for teaching industrial robotics remotely. 
    \item Study our platforms' effectiveness by deploying them in a PBL-based competition spread over six months with more than 1000 students.
\end{enumerate}


\section{Methodology}
\label{sec:methodology}

\subsection{Setup of Platforms}
\label{sec:meth_intro}
Scalability is one of the main factors considered while designing our platforms. While adding multiple work cells can easily increase the number of students that access a remote laboratory, it comes at an additional cost. Even though the simulation created, as shown in Figure~\ref{fig:aahs_awms_intro}, for each platform closely mimicked the real-world, we can not eliminate the sim-to-real gap. Hence, in both of our platforms, students first tested their algorithm using a simulator, Gazebo~\cite{gazebo}, and then only selected students (50-100), based on their performance, tested them using remote hardware. The students connected to the hardware in one-hour slots for three months. These students participated by forming a team of 2-4 members in our 7-month eYRC competition.

The multi-robot system (MRS) used in both of our platforms is shown in Figure~\ref{fig:aahs_awms_intro}. The four-wheeled mobile rover works on a skid steer drive control mechanism. It is powered by Intel NUC12WSHI7 8GB and comprises various sensors such as encoders, 2D LIDAR, and an Inertial Measurement Unit (IMU). The UR5 robotic arm~\cite{ur5} has a payload capacity of 5 kg and a reach of 850 mm. Custom interchangeable grippers, like a parallel arm or magnetic, can be interfaced easily with the arm.


\subsection{Stack 1: Using peer-to-peer VPN only}
\label{sec:stack1}
\begin{figure}[!t]
    \centering
    \includegraphics[width=1\linewidth]{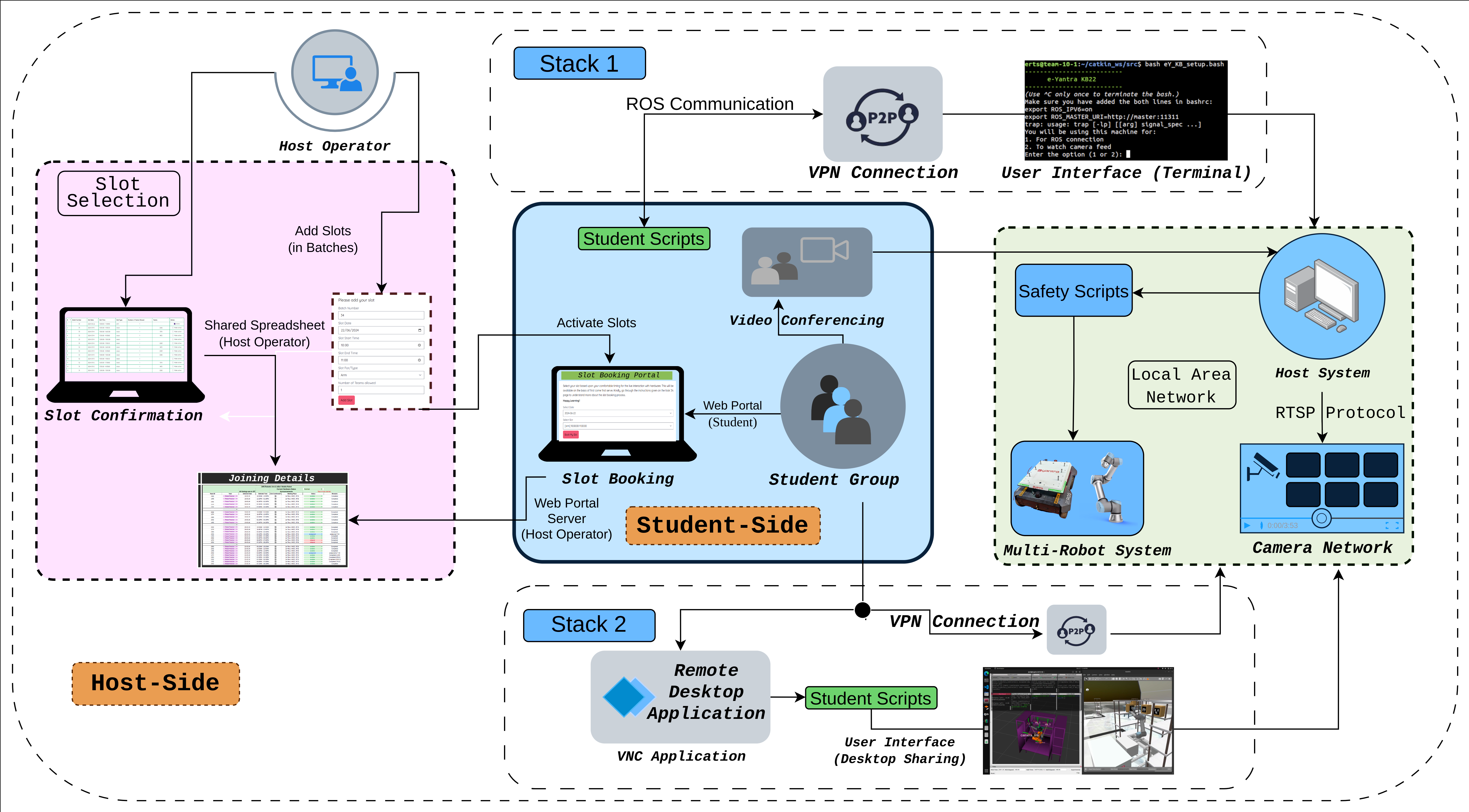}
    \caption{Operational Workflow Architecture of Stacks 1 and 2 for remote lab platform: The architecture is divided into two major sections: The Student and Host Side. Stack 1 uses only a peer-to-peer VPN, whereas Stack 2 uses a combination of peer-to-peer VPN and remote desktop application.}
    \label{fig:common_architecture}
\end{figure}

We propose a peer-to-peer VPN-based architecture for remote operation using the ROS framework, as depicted in Figure~\ref{fig:common_architecture}. The mobile rover and the UR5 arm traverse inside a greenhouse at IIT Bombay with artificial plants, as shown in Figure~\ref{fig:aahs_awms_intro}(top-right). The students have to pluck artificial fruits (attached to plants with a magnet) using a gripper connected to the arm and deposit them in a box in front of the robot. A student-side computer uses ROS communication protocols in this stack to control the MRS, including navigation, manipulation, and perception. A peer-to-peer VPN like Husarnet \cite{husarnet} connection allows students to run their code directly on the remote hardware using their workspace. Consequently, only data is exchanged between the MRS and the student's computer, creating a virtual ecosystem for students to work on the MRS. For visual feedback, they are provided with a webpage showing all the live camera feeds using the RTSP \cite{rtsp} protocol.

The sub-system of the architecture, as shown in Figure~\ref{fig:common_architecture}, is divided into two sections: student and host side. Detailed explanations of the same are given below: 

\begin{enumerate}
    \item \textbf{Student-Side:}
        \begin{enumerate}
            \item \textbf{Slot selection:} The students are made aware of the upcoming remote hardware access slots using a common discussion forum. They can book their slots by filling out a form. Once the slot is booked, the same is reflected on a shared spreadsheet showing the time, team-id, status of the slot, and a joining link to an online video call platform.
            
            \item \textbf{VPN Host:} A bash script \cite{bash} is used to connect the student's computer to the host computer network. This is done by using an open-source peer-to-peer VPN. A joining code to join the VPN is given by the host. The student can publish or subscribe to all the data with MRS using the ROS communication protocols once they get connected. The average latency for data communication in this stack has been observed to be 300 milliseconds. The student is provided with the IPV6 (Internet Protocol Version 6) link and a port number for the web page hosting the live camera feed using RTSP \cite{rtsp} protocol. A lag of approximately 2 seconds has been observed at the student's end in viewing the camera feed. This does not affect the control of the MRS as communication with MRS is separate.
            
        \end{enumerate}

    \item \textbf{Host-Side:}
        \begin{enumerate}
        \item \textbf{Communication and Remote Access:} The host and student are in the same network using the peer-to-peer VPN. This allows for continuous monitoring of the data that is transferred to the MRS. The names and types of the topics in ROS are kept the same as in the simulation. Thus, with negligible changes, students can run their algorithm in remote hardware that was developed for simulation.
        \item \textbf{Host System:} This physical system consists of a computer with an open-source Ubuntu \cite{sobell2015practical} OS connected to the MRS in a local wireless network. This system has all the drivers installed to relay data between the student and the MRS. The data is mapped to different ROS topic names in the host computer using a Python \cite{python} script. This allows the host to have control of MRS in case of emergency or need. This system is also the server for hosting 5 PoE (Power over Ethernet) enabled CCTV (Closed Circuit Television) cameras on a webpage using RTSP protocol.
        \item \textbf{Multi-Robot System:} The MRS has the UR5 mounted on the top of the mobile rover. Using USB (Universal Serial Bus) and ethernet cable, the depth camera and UR5 are connected respectively to the mobile rover's computer. Since the host computer and the MRS are in the same local network, the host operator can use SSH (Secure Shell) to access files or run any program on the MRS.
        \end{enumerate}

\end{enumerate}




\subsubsection{Safety Measures:}
The safety and security of stack 1 also come into play since the architecture involves remote access of MRS. Any mishaps due to the collision of MRS with walls or objects can happen due to incorrect values of navigation or manipulation commands sent by students. The latency in communication can also play a significant role here. To avoid any mishap, the host operator is notified about a possible collision by a) using a local Python script running in the MRS that checks for collision by measuring distance using LIDAR (Light Detection And Ranging) sensor values, b) continuously checking the frequency of incoming data from MRS on the host system, c) adding a safety stop script for UR5 in the MRS to stop when a collision is detected while manipulation, d) adding speed, joint and movement-space limits to the UR5 arm. The host system stops the relay of data between MRS and the student if it detects any of the points mentioned above crossing a specific threshold. The operator is also notified using different sounds from the host system for added safety.


\subsection{Stack 2: Using remote desktop and VPN}
\label{sec:stack2}
In this stack, we propose a remote desktop-based architecture using the ROS 2~\cite{ros2Paper} framework to facilitate monitoring and remote control of AWMS as shown in Figure~\ref{fig:aahs_awms_intro}. Students use this system to autonomously navigate a mobile robot to collect racks in the warehouse and deliver them to the robotic arm using the ROS 2 navigation stack ~\cite{navigationStack}. They then sort packages on the rack using camera-based perception. Hence, students can work on the mobile rover for mapping and navigation and on the robotic arm for perception and manipulation, individually and in tandem, to perform warehouse automation tasks. Installing ROS 2~\cite{ros2-doc-humble} along with the necessary packages is time-consuming and may dispirit new learners. Additionally, executing the ROS 2 scripts within the local network ensures low latency and high repeatability. Hence, our approach is to give students direct control of the host system's desktop. A student only needs a laptop with any operating system installed. Figure~\ref{fig:common_architecture} demonstrates this stack's architecture. The architecture is divided into two sub-systems: the Student-Side and the Host-Side.


\begin{enumerate}
    \item \textbf{Student-Side:}
        \begin{enumerate}
            \item \textbf{Slot selection}: The process for slot selection is similar to Stack 1, as mentioned above. To enhance user interface and provide seamless experience, a web portal was designed for booking slots using an open-source tech stack, namely the Vue Inertia Laravel Tailwind (VILT). The host operator can monitor all slots that have been created and have access to activate (add the slot for students) or deactivate (disable the slot for students) any slot as per lab requirements. Slots that the host operator has confirmed are communicated to the student teams. A spreadsheet is generated from the portal and is shared with all students. The student connects to the host-side operator through video conferencing for setup and coordination.
            \item \textbf{VPN Host:} Same as stack 1, but instead of using bash script, we directly used the video call communication for the camera feed IPV6 link, simplifying the setup.
            
        \end{enumerate}
    \item \textbf{Host-Side:}
        \begin{enumerate}
            \item \textbf{Communication and Remote Access:} The stack uses a free and platform-independent Virtual Network Computing (VNC) application, like Anydesk~\cite{anydesk}, for remote desktop sharing. It provides students access to the host system for remote operations.
            \item \textbf{Host System:} It consists of a computer with an open-source Ubuntu~\cite{sobell2015practical} operating system installed. It has all the software/hardware drivers, ROS 2~\cite{ros2Paper} packages, and RViz~\cite{rviz} installed. This host system serves the following purposes - host desktop for VNC application and host device for camera setup. The UR5 robotic arm and the mobile rover are also controlled using Python scripts on this host system.
            \item \textbf{Multi-Robot System:} This system consists of two separate robots; the mobile rover is connected via Wi-Fi to execute the navigation commands, while the UR5 robotic arm and a depth camera are connected to the host system using an Ethernet \& USB cable for manipulation and perception. Since both robots operate in the same VPN, the students can access the robots' data on the host system and on their laptops.
        \end{enumerate}
\end{enumerate}





\subsubsection{Safety Measures:}
We incorporated a few strategies to reduce privacy and security threats related to remote host access. To prevent unauthorized access during remote sessions, we restricted student access to their designated ROS workspace only. For this, we used the Encoded File System (EncFS)~\cite{encfs} package to encrypt personal directories for security. We also adopt a non-privileged user account strategy to protect the integrity of the host system. This dramatically improves security by stopping students from unintentionally or intentionally erasing files, altering system settings, or installing unwanted software. Students don't need superuser access as the necessary software packages are pre-installed. Lastly, the VNC software allows file transferring and other remote access features. Hence, we use VNC software like Anydesk to control and monitor sessions, which provides admin privileges. The host can revoke desktop access, reset IP addresses, control file transferring, and manage session access. Additionally, we use safety scripts to prevent robot collisions during student testing, as in Stack 1.

\section{Results}
\label{sec:results}
\begin{figure}[!t]
\begin{subfigure}{.5\textwidth}
  \centering
  \includegraphics[width=1\linewidth]{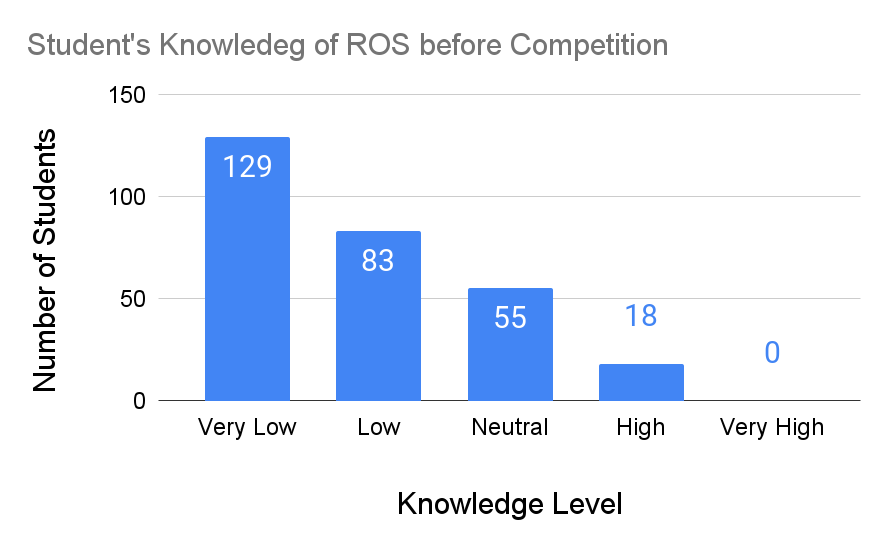}
  \caption{eYRC 2022-23}
  \label{fig:Self-Reported-2022}
\end{subfigure}%
\begin{subfigure}{.5\textwidth}
  \centering
  \includegraphics[width=1\linewidth]{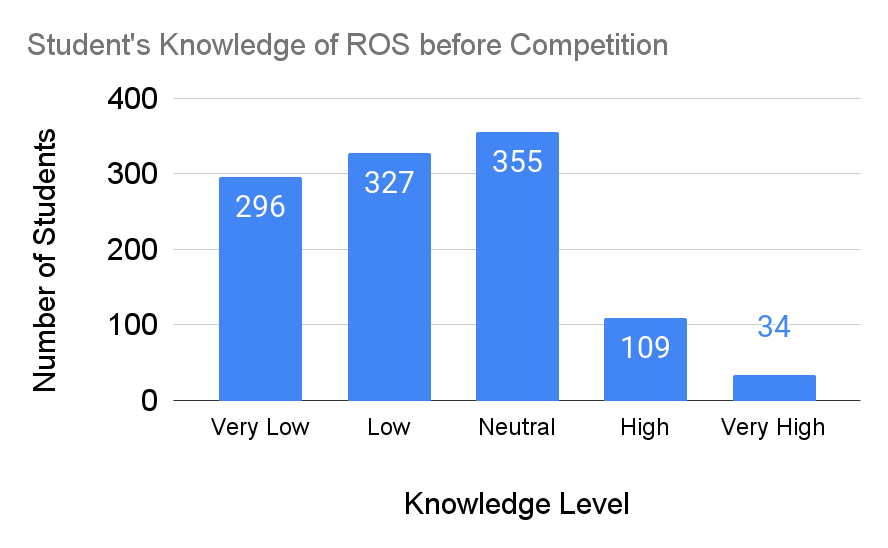}
  \caption{eYRC 2023-24}
  \label{fig:Self-Reported-2023}
\end{subfigure}
\caption{Self-reported expertise by students in both years of eYRC before the start of the competition.}
\label{fig:Self-Reported}
\end{figure}

We developed problem statements for eYRC 2022-23 and 2023-24 based on the proposed platforms. These problem statements were divided into tasks, each spanning approximately three weeks, and completing each helped the students develop the entire solution. We provided learning resources and set up online discussion forums. The competition was divided into two stages: stage 1 for simulation and stage 2 for hardware implementation. The teams were continuously evaluated based on their relative performance in these tasks. The top-performing teams were selected to proceed to the hardware implementation stage.


\subsubsection{eYRC 2022-23(AAHS):}
Stack 1 was used for the eYRC competition of 2022-23. The task for the students was to develop algorithms for the AAHS, set up in a greenhouse at IIT Bombay. The competition here was divided into six tasks. The first four tasks were simulation-based, whereas we provided remote access for hardware implementation during tasks 5 and 6.
A total of  373 teams, with 1433 students, registered for the competition. Out of which 197 participating teams submitted the first task, i.e., task 0. We asked students to self-rate their knowledge of ROS at the start of the competition, as shown in Figure \ref{fig:Self-Reported-2022}.

\begin{figure}[!t]
\begin{subfigure}{.5\textwidth}
  \centering
  \includegraphics[width=1\linewidth]{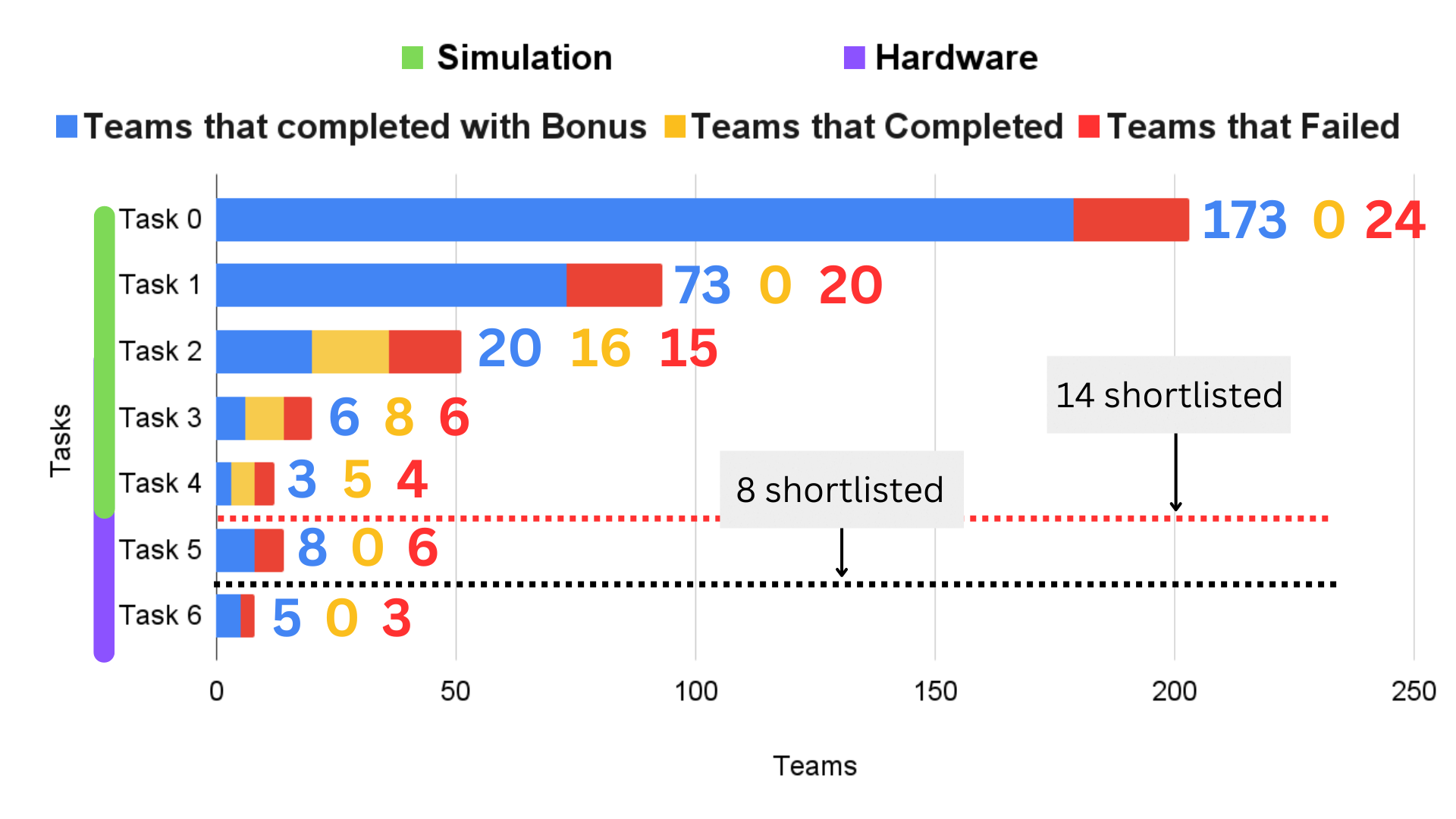}
  \caption{eYRC 2022-23}
  \label{fig:scores-2022}
\end{subfigure}%
\begin{subfigure}{.5\textwidth}
  \centering
  \includegraphics[width=1\linewidth]{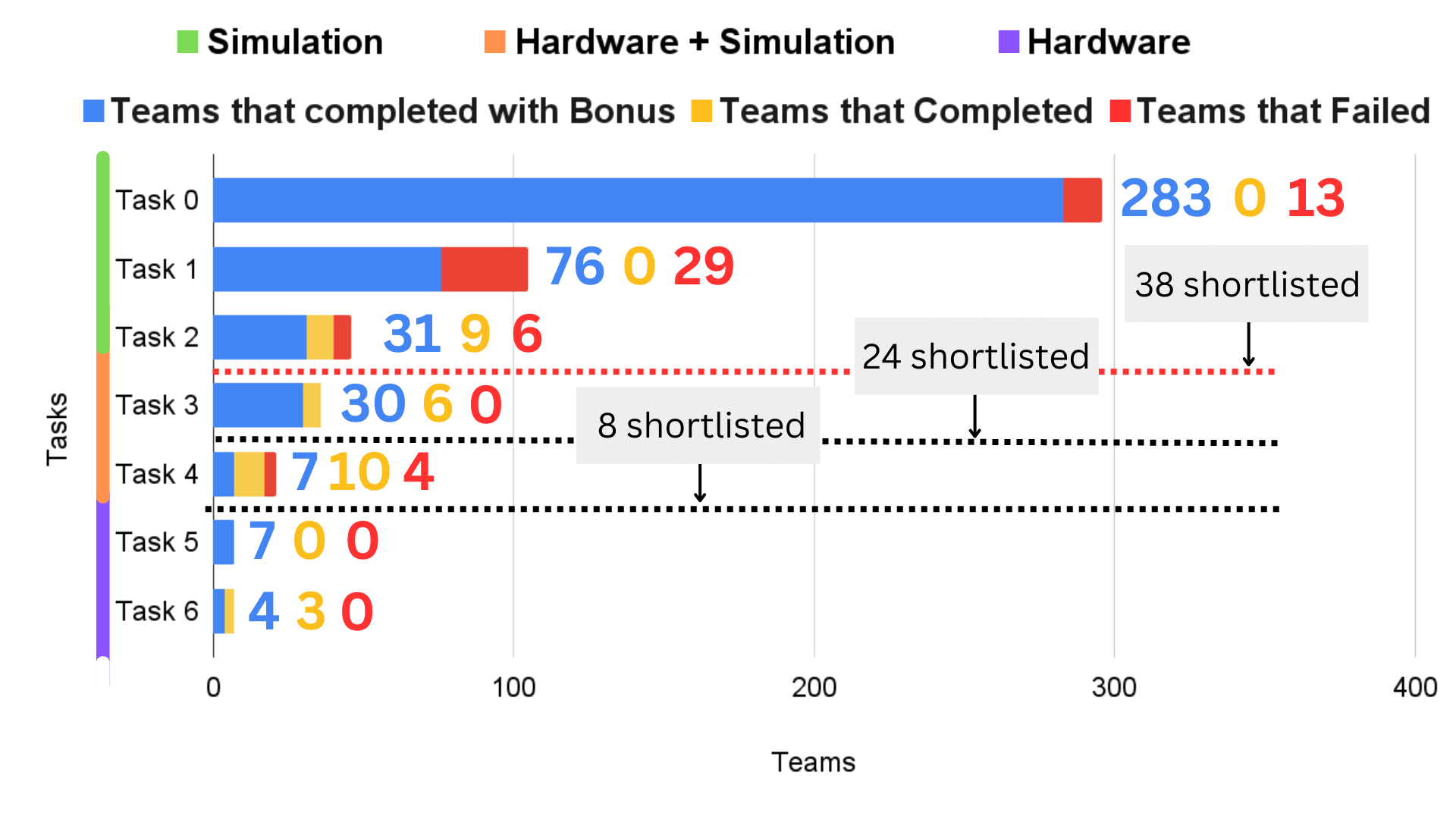}
  \caption{eYRC 2023-24}
  \label{fig:scores-2023}
\end{subfigure}
\caption{Performance of teams in each task during the two separate editions of eYRC. A total of 197 teams participated in eYRC 2022-23, whereas 296 teams participated in eYRC 2023-24. Teams were awarded a bonus if they completed the submission within the deadline.}
\label{fig:scores}
\end{figure}

The first task, i.e., task 0, required students to set up the ROS workspace and development environment. The task was to draw a shape using Turtlesim. They were followed by tasks 1, 2, and 3, where we taught students about the navigation of mobile rover, perception, and manipulation of UR5 Arm in the Gazebo simulation environment. For task 4, the students had to develop robust algorithms to completely automate the AAHS in the shortest time possible in simulation. Figure \ref{fig:scores-2022} shows the number of teams that were able to complete the given tasks. The data signifies that this approach was successful in educating 173 teams on the basics of ROS. The participating teams, primarily consisting of beginners in robotics, developed the algorithm for AAHS in simulation till task 4.


A known issue in MOOCs has been the high dropout rates \cite{doi:10.1080/15313220.2020.1809050}. We faced the same problem, with 54 percent of teams dropping out after task 0. After this, we experienced around 45 percent dropouts as the complexity increased with each task, as shown in Figure \ref{fig:scores-2022}.  We had no dropouts after task 4.
Based on the teams' performance in all four simulation tasks, 14 teams were shortlisted for the hardware implementation. To ease the sim-to-real transition, we divided the hardware tasks into sub-tasks, as shown in Figure \ref{fig:slots-2022}. Although we didn't have teams dropping out of the competition altogether after task 4, some of the participating teams missed some remote access slots, which can be observed in Figure \ref{fig:slots-2022}. We progressively increased the complexity of these tasks, starting with perception using a depth camera, followed by manipulation of the UR5, and finally, the navigation of the mobile rover. As complexity increased, more remote access slots were provided. Stack 1 was tested over 160 slots of 1 hour each by 70 students over three months to remotely access the robots.


\subsubsection{eYRC 2023-24(AWMS):}
Stack 2 was implemented during the eYRC 2023-24. The task for the students was to develop an algorithm for the AWMS that was set up in a warehouse at IIT Bombay. A total of 1312 students forming 349 teams registered for the competition. Out of which 296 teams completed the task 0. The simulation stage here was divided into two main tasks and five sub-tasks. We asked students to self-rate their knowledge of ROS at the start of the competition, as shown in Figure \ref{fig:Self-Reported-2023}. Task 0 was very preliminary and only included software installation and workspace setup. For tasks 1 and 2, students had to study image processing and develop a script to manipulate the UR5 ARM to pick up the payload (boxes). Students also had to complete autonomous navigation using the mobile rover in the simulation environment. The performance of all participating teams can be seen in Figure \ref{fig:scores-2023}.

\begin{figure}[!t]
\begin{subfigure}{.5\textwidth}
  \centering
  \includegraphics[width=1\linewidth]{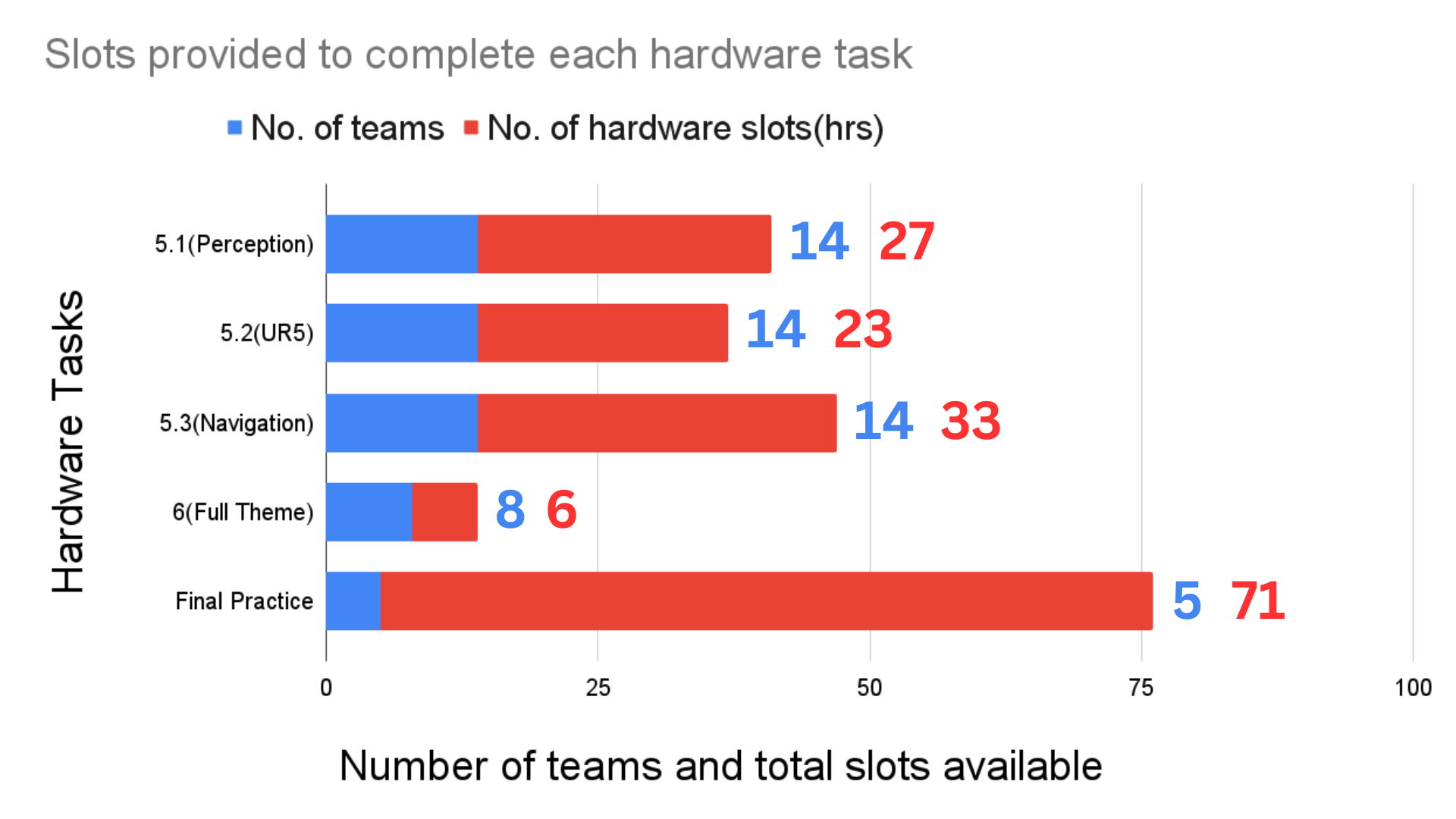}
  \caption{eYRC 2022-23}
  \label{fig:slots-2022}
\end{subfigure}%
\begin{subfigure}{.5\textwidth}
  \centering
  \includegraphics[width=1\linewidth]{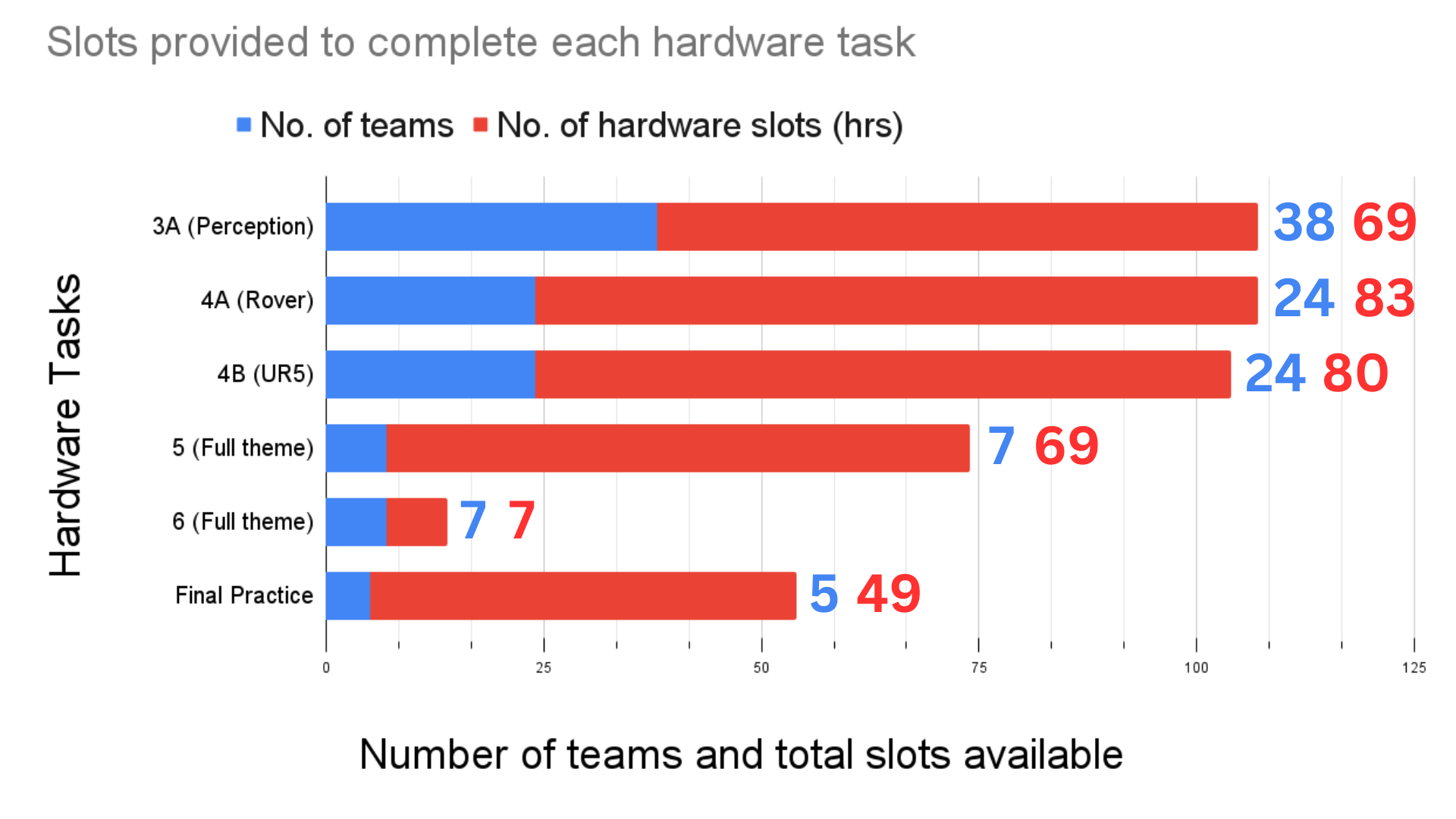}
  \caption{eYRC 2023-24}
  \label{fig:slots-2023}
\end{subfigure}
\caption{Total remote access slots (hours) given per hardware task vs the number of teams participating in the task. More remote access hours were provided for complex tasks.}
\label{fig:slots}
\end{figure}

In 2023-24, we shortlisted teams in three stages, as shown in Figure \ref{fig:scores-2023}. First, based on the performance in the simulation stage, we shortlisted 38 teams for stage 2, i.e., hardware implementation. The following two short listings were based on the team's relative performance in the remote access hardware slots. For task 4, 24 teams were shortlisted, of which only four teams dropped out. Here, we had a mixture of both simulation and hardware. The students were provided 4 hours of remote access to the robots, in which they had to pick and place boxes using UR5 and implement navigation and docking of the mobile rover. As the last sub-task of task 4, teams had to complete the entire task implementation in simulation. The teams were graded relatively. Similar to stack 1, we implemented the strategy to introduce each hardware component as a small sub-task, as shown in Figure \ref{fig:slots-2023}. This strategy helped students easily overcome the sim-to-real gap and quickly tackle complex automation problems.

For stack 2, 76 teams consisting of beginners were able to complete both mobile rover navigation and arm manipulation using image processing in the simulation stage. All 36 teams that got remote access achieved the results on hardware, as shown in Figure \ref{fig:scores-2023}. During eYRC 2023-24, we successfully provided remote access to 36 teams, i.e., 146 students. Stack 2 was tested for over 355 hours of remote access. We successfully trained 1312 students in robotics and ROS during the eYRC 2023-24 using stack 2. Although the MOOC dropout rate has been a prominent issue, after the initial drop of more than 50 percent, we retained a significantly high number of participants during stage 2, with only 6 teams dropping out, as shown in Figure \ref{fig:scores-2023}.

Table~\ref{tab:stack_1_2_comp} highlights the key differentiating factors of both the approaches demonstrated by stack 1 and stack 2.
The finalists of the competition in 2022-23 got an average of 19.4 hours of remote access to the robots, while those of eYRC 2023-24 got an average of 30.8 hours. This, combined with the data from Figure \ref{fig:scores}, signifies that finalists, primarily consisting of beginners, could completely automate AAHS and AWMS in such limited hours. 

\begin{table}[h!]
\setlength{\arrayrulewidth}{0.3mm}
\setlength{\tabcolsep}{5pt}
\renewcommand{\arraystretch}{1.5}
\centering
\caption{Key differentiating factors of stack 1 and stack 2, based on the data acquired from eYRC 2022-23 and eYRC 2023-24.}
\label{tab:stack_1_2_comp}
\scriptsize
\begin{tabular}{ |p{2.7cm}|p{4cm}|p{4.1cm}|}
\hline
\centering  \cellcolor[HTML]{FFF2CC} - & \centering \cellcolor[HTML]{FFF2CC} Stack 1 & \centering \cellcolor[HTML]{FFF2CC}  Stack 2 \tabularnewline \hline
\centering Medium & \centering Peer-to-Peer VPN & VNC Graphical and Peer-to-Peer VPN \\ \hline
\centering Type of Robot(s)  & \multicolumn{2}{p{8cm}|}{	\centering Static and mobile together in single testbed} \\ \hline
Testbed used in Remote Lab (m- meters)  &	Greenhouse Size: 10m x 6m   & Warehouse Size: 8m x 6m \\ \hline
Algorithm Execution & \centering Student-Side  & \centering Host-Side \tabularnewline \hline
Communication &	\multicolumn{2}{p{8cm}|}{The student, host, and robots are in the same VPN. ROS is used for bidirectional data communication.}	 \\ \hline
No. of student teams (trained on hardware) & \centering 14 &  \centering 38 \tabularnewline  \hline
Hardware access given to students (over 3 months) & \centering \vspace{+0.2mm} 160 hours & \centering \vspace{+0.2mm} 355 hours \tabularnewline \hline
\centering \vspace{+5.8mm} Advantages & 	
\vspace{-4mm}
\begin{itemize}
    \item Simple host-side setup
    \item Students have the freedom to use any packages/dependencies.
    \item Safety scripts run on the host-side to prevent damage to the robots and the arena.
\end{itemize} 
&
\vspace{-4mm}
\begin{itemize}
\item Low latency during execution of algorithm
\item Low minimum system requirements for students
\item Safety scripts run on the host-side to prevent damage to the robots and the arena.
\end{itemize}
\\ \hline

\centering \vspace{+5.8mm} Disadvantages &
\vspace{-4mm}
\begin{itemize}
    \item Average latency of 300 ms
    \item High system requirements for students developing computationally demanding algorithms.
\end{itemize}
&
\vspace{-4mm}
\begin{itemize}
\item Complicated host-side setup
\item High setup time: Students need to copy their scripts to the host pc at the beginning.
\item Students are restricted to using packages supported by the host.
\end{itemize}
\\ \hline
\end{tabular}
\end{table}

\section{Conclusion}
\label{sec:conclusion}
In this paper, we present two remote platforms for teaching industrial robotics to students. Compared to previous work, our platforms use open-source tools and are tested with thousands of students. Testing these platforms as a part of our robotics competition, eYRC, showed that the finalist teams could develop fully autonomous algorithms for robots in a greenhouse and a warehouse with an average of 19.4 hours and 30.8 hours of remote hardware access, respectively. In eYRC 2022-23, we provided training to 197 teams using a simulator and 14 teams using hardware. For eYRC 2023-24, this increased to 296 teams trained on the simulator and 38 teams trained with hardware. A comparison of both approaches has been presented with its advantages and disadvantages. One major drawback of a remote lab in a full-size real-world setting is that the host operator has to reset the components after every run. This reduces the hours a remote lab can be operated each day. Additionally, the students should be able to automatically reset the robots and the various components in the arena. Future work will focus on automating the greenhouse and the warehouse to minimize host operator involvement, improving the safety of the robots, reducing the latency between the host and the student's system, and maximizing the remote lab's efficiency in terms of lab hours.

\vspace{-0.15cm}
\begin{credits}
\section{\ackname} We would like to thank the Ministry of Education (MoE), Govt. of India, for their support. This work was funded by MoE, Govt. of India, in the project e-Yantra (RD/0121-MHRD000-002). We are grateful to Arjun Sadananda, Ravikumar Chaurasia, Saail Narvekar, Simranjeet Singh, Smit Kesaria, Soofiyaan Atar, and Vishal Gupta for their support. This project utilized open-source software packages, too many to enumerate. We sincerely thank all the software developers and users for their contributions. We also thank anonymous reviewers for their insightful comments and suggestions.

\subsubsection{\discintname}
The authors have no competing interests to declare that are relevant to the content of this article.
\end{credits}
%
%
%
%
\bibliographystyle{splncs04}
{\small
\bibliography{references}}

\end{document}